\documentclass[letterpaper, 10 pt, conference]{ieeeconf}  

\IEEEoverridecommandlockouts                              

\usepackage{amsmath,amsfonts}
\usepackage{algorithmicx}
\usepackage{algorithm}
\usepackage{array}
\usepackage[caption=false,font=normalsize,labelfont=sf,textfont=sf]{subfig}
\usepackage{textcomp}
\usepackage{stfloats}
\usepackage{url}
\usepackage{verbatim}
\usepackage{graphicx}
\usepackage{cite}
\usepackage{amsmath,amssymb,amsfonts}
\usepackage{xcolor}
\usepackage{hyperref}
\usepackage{svg}
\usepackage{multirow}
\usepackage{subcaption}
\usepackage{algpseudocode}
\usepackage{CJKutf8}
\usepackage{enumerate}
\usepackage{makecell}
\usepackage{fancyhdr} 
\usepackage{booktabs}
\usepackage{tabularx}

\title{\LARGE \bf
Embodied Intelligence in Disassembly: Multimodal Perception Cross-validation and Continual Learning in Neuro-Symbolic TAMP
}

\author{Ziwen He$^{1}$, Zhigang Wang$^{2}$, Yanlong Peng$^{1}$, Pengxu Chang$^{1}$, Hong Yang$^{3}$, Ming Chen$^{1}$*
\thanks{This work was supported by the Ministry of Industry and Information Technology of China for financing this research within the program "2021 High Quality Development Project (TC210H02C)"}
\thanks{$^{1}$ School of Mechanical Engineering, Shanghai Jiao Tong University, Shanghai, China. (e-mail: \{heziwen, me-pengyanlong, 19119175490, mingchen\}@sjtu.edu.cn)}%
\thanks{$^{2}$ Intel Labs China, Beijing, China. (e-mail: zhi.gang.wang@intel.com)}%
\thanks{$^{3}$ Intel Asia Pacific R\&D Ltd, Shanghai, China. (e-mail: harold.yang@intel.com)}%
}
\begin{document}

\maketitle
\thispagestyle{empty}
\pagestyle{empty}

\begin{CJK*}{UTF8}{gbsn} 
\begin{abstract}
With the rapid development of the new energy vehicle industry, the efficient disassembly and recycling of power batteries have become a critical challenge for the circular economy. In current unstructured disassembly scenarios, the dynamic nature of the environment severely limits the robustness of robotic perception, posing a significant barrier to autonomous disassembly in industrial applications. This paper proposes a continual learning framework based on Neuro-Symbolic task and motion planning (TAMP) to enhance the adaptability of embodied intelligence systems in dynamic environments. Our approach integrates a multimodal perception cross-validation mechanism into a bidirectional reasoning flow: the forward working flow dynamically refines and optimizes action strategies, while the backward learning flow autonomously collects effective data from historical task executions to facilitate continual system learning, enabling self-optimization. Experimental results show that the proposed framework improves the task success rate in dynamic disassembly scenarios from 81.68\% to 100\%, while reducing the average number of perception misjudgments from 3.389 to 1.128. This research provides a new paradigm for enhancing the robustness and adaptability of embodied intelligence in complex industrial environments.
\end{abstract}

\section{INTRODUCTION}

\begin{figure*}[t]
  \centering
  \includegraphics[width=\linewidth]{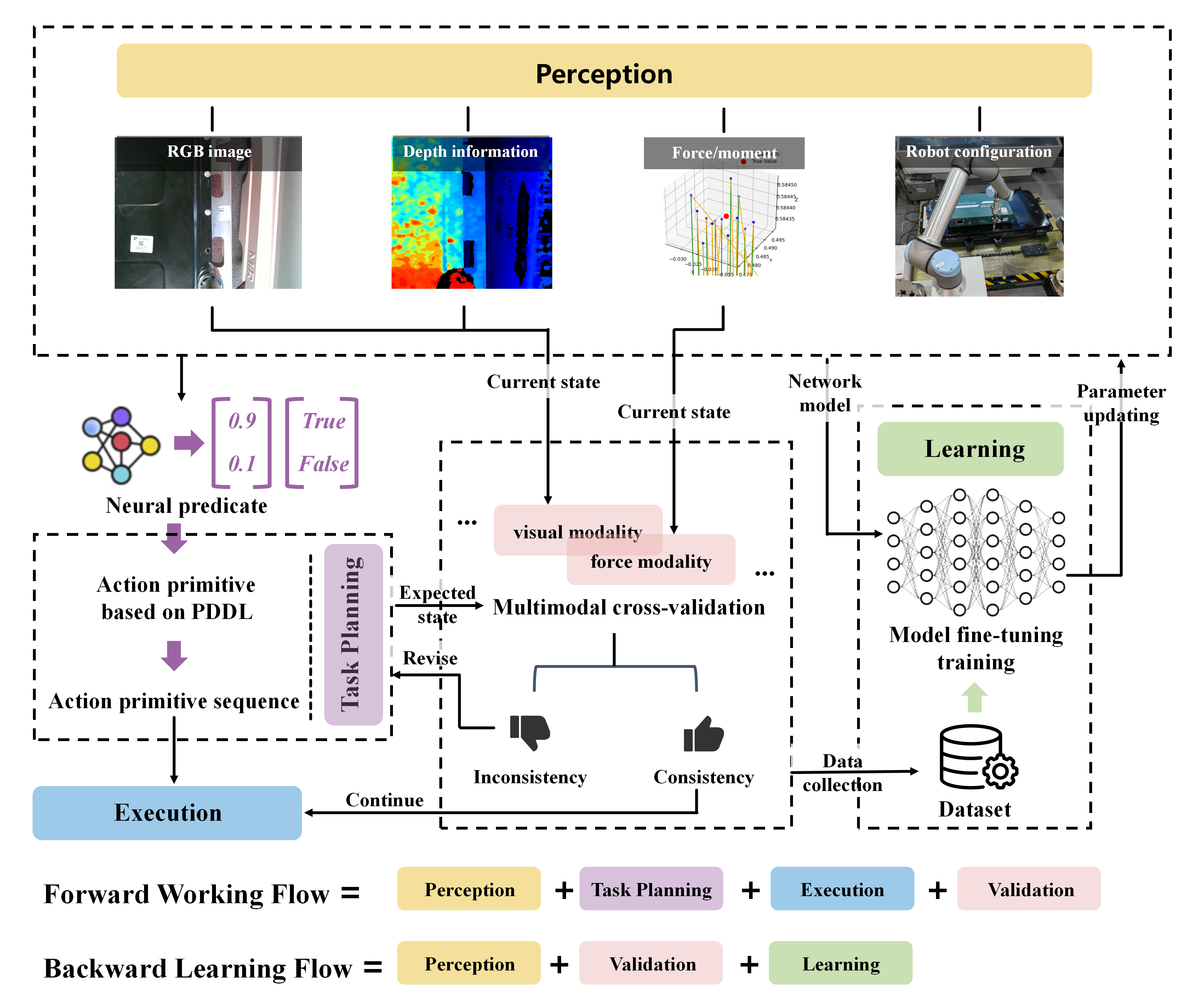}
  \caption{System framework diagram. The system operates under a closed-loop control architecture composed of five modules: perception, planning, execution, verification, and learning. In the forward working flow, the perception module updates neural predicates based on environmental inputs, the planning module generates action planning sequences, and the execution module carries out these sequences. During task execution, the verification module dynamically adjusts task strategies through vision-force cross-validation. In the backward learning flow, the learning module infers from historical records in the verification module, automatically collecting valid data through the cross-validation mechanism to expand the continual learning dataset and optimize the performance of the perception module.}
  \label{fig:framework}
\end{figure*}

With the rapid development of Industry 4.0 and the circular economy, industrial disassembly has become a critical link in intelligent manufacturing and resource recycling, facing unprecedented technical challenges~\cite{kay2022robotic,baazouzi2021optimization}. Traditional disassembly operations heavily rely on manual labor, resulting in low efficiency, high costs, and poor safety. Embodied intelligence, by endowing robots with perception, decision-making, and execution capabilities, offers a new technological pathway toward autonomous disassembly. However, the complexity of industrial disassembly scenarios imposes extremely high demands on robots' multimodal perception, task planning, and continual learning capabilities.

Task and motion planning (TAMP), as one of the core technologies of embodied intelligence, achieves closed-loop adaptability in complex scenarios by integrating symbolic task planning based on the Planning Domain Definition Language (PDDL)~\cite{aeronautiques1998pddl,fox2003pddl2,younes2004ppddl1} with continuous-space motion optimization through a hierarchical architecture~\cite{garrett2021integrated}. In recent years, Neuro-Symbolic approaches, which combine the perceptual learning capabilities of neural networks with the logical reasoning power of symbolic systems, have shown great potential in the TAMP domain~\cite{shen2018neuro}.

In our previous work, we designed a Neuro-Symbolic autonomous disassembly system for robots~\cite{zhang2022autonomous,zhang2023design,zhang2023development}. This system dynamically maps multimodal perception to symbolic states through neural predicates, transforming sensory data — such as visual image analysis and force feedback — into discrete symbolic representations. Based on a hierarchical task planning framework, the system autonomously schedules action primitives. The system successfully achieved autonomous disassembly of electric vehicle battery screws and demonstrated good generalization ability in unstructured environments.

However, the existing system struggles to adapt to dynamic changes during operation. First, environmental disturbances can induce perception shifts. Layout changes (e.g., camera displacement or sudden lighting variations) can cause a distribution shift in perception features, leading to symbolic state misjudgments and action sequence conflicts. Second, long-term operational degradation occurs. In continuous disassembly tasks, tool wear gradually causes end-effector positioning drift, destabilizing the perception-planning-execution loop. The current system relies on fixed-parameter models, making it difficult to dynamically adapt to these time-varying disturbances, ultimately limiting the system's robustness for industrial-grade applications in unstructured disassembly tasks.

To address these challenges, this study proposes an adaptive optimization framework that integrates multimodal cross-validation and continual learning (Figure~\ref{fig:framework}). The system tightly couples inference and decision-making with perception and control, embedding the vision-force cross-validation mechanism into the PDDL-based forward working flow and backward learning flow. This framework dynamically refines strategies during task execution and effectively incorporates historical data into continual learning, constructing a closed-loop control system of perception-planning-execution-validation-learning.

Our main contributions are as follows:
\begin{itemize}
\item We propose a multimodal cross-validation mechanism that enhances embodied intelligence adaptation in dynamic environments through vision-force perception cross-validation and neural predicate dynamic correction.
\item We design a self-optimizing continual learning framework, introducing a PDDL-based bidirectional reasoning flow with forward task execution and backward learning optimization. The system automatically constructs training datasets from historical robot tasks, reducing dependence on manual annotations and achieving continual model parameter optimization.
\item We conduct experiments in a real-world battery disassembly scenario to validate the proposed continual learning framework. The results show that the screw disassembly success rate increases from 81.68\% to 100\% under environmental disturbances, while the average number of perception misjudgments per task decreases from 3.389 to 1.128. These results demonstrate the robustness and practicality of the framework in unstructured dynamic environments, providing a technical paradigm for the industrial deployment of embodied intelligence.
\end{itemize}

\section{RELATED WORK}
\subsection{Neuro-Symbolic TAMP}
Artificial intelligence has evolved from symbolic reasoning to deep learning-based probabilistic approaches. While symbolic methods offer strong interpretability, they struggle with complex and dynamic environments. In contrast, deep learning excels at perception tasks but relies heavily on large-scale data and lacks reasoning capabilities. Neuro-Symbolic AI, which integrates the strengths of both paradigms, has emerged as a promising research direction~\cite{susskind2021neuro,hitzler2022neuro}.

Neuro-Symbolic TAMP is a representative application of this paradigm. It uses neural predicates to model environment states and action primitives to bridge high-level task descriptions with low-level motion execution, enabling robust planning in unstructured environments~\cite{shen2018neuro,zhang2023novel,zhang2023development}.

Despite improvements in flexibility and efficiency, current Neuro-Symbolic TAMP systems still face challenges, such as perception errors affecting symbolic decision-making and limited adaptability to dynamic scenarios. A key issue is how to effectively integrate multimodal perception and adopt continual learning to dynamically refine neural-symbolic mappings, thereby enhancing system robustness.

\subsection{Continual Learning and Autonomous Data Acquisition}
Continual learning (CL) enables robots to adapt to evolving tasks in dynamic environments while retaining previously acquired knowledge and mitigating catastrophic forgetting~\cite{hadsell2020embracing, pfulb2018catastrophic}. Current CL strategies encompass a variety of approaches, including gradient-based regularization~\cite{clopath2008tag}, modular decomposition~\cite{aljundi2018selfless}, memory-augmented methods~\cite{chaudhry2021using}, and meta-learning~\cite{triantafillou2019meta}. These techniques demonstrate different strengths in interference mitigation, parameter isolation, and fast adaptation. Increasingly, they are being integrated with deep neural networks to form diverse and robust CL frameworks.

In terms of data acquisition, weakly-supervised learning has emerged as a promising solution to reduce reliance on manual annotations. For example, Barnes et al. proposed a weakly-supervised segmentation method for autonomous driving that leverages vehicle trajectories to automatically generate large-scale labeled datasets for path and obstacle detection~\cite{barnes2017find}. Inspired by such approaches, our work adopts a task-driven data acquisition paradigm, where effective training data is collected based on the robot's own behavior and perception, facilitating self-supervised optimization without human labeling.

\subsection{Continual Learning in Neuro-Symbolic TAMP}
Recent research has introduced continual learning into Neuro-Symbolic TAMP frameworks to enhance adaptability in open and dynamic environments. For instance, HyGOAL~\cite{lorang2024framework} integrates symbolic planning with reinforcement learning to enable robots to quickly adapt to novel situations such as environmental disturbances and primitive failures. However, existing approaches primarily focus on strategy-level generalization and often lack mechanisms to support continual learning in the perceptual components. 

To address these limitations, our proposed Neuro-Symbolic TAMP system incorporates a multimodal cross-validation mechanism based on visual and force feedback. This allows the system to continuously refine both its symbolic state representations and perceptual models without requiring external annotations. By bridging the gap between neural perception and symbolic reasoning, our approach improves planning robustness and environmental adaptability in complex robotic tasks.

\section{PROBLEM DEFINITION} 

\begin{table}[t]
  \caption{Detailed Description of (Neural) Predicates}
  \label{tab:predicates_description}
  \begin{center}
  \begin{tabularx}{\linewidth}{cX} 
  \toprule  
  Predicate & \multicolumn{1}{c}{Function Description} \\
  \midrule  
  \texttt{have\_coarse\_pose} & The robot has obtained the coarse position of the screw. \\
  \texttt{pattern} & 1 indicates visual perception, 0 indicates force perception. \\
  \texttt{near\_screw} & The end-effector is close to the screw. \\
  \texttt{above\_screw} & The end-effector is near the top of the screw. \\
  \texttt{disassembled} & The screw has been disassembled. \\
  \texttt{target\_aim} & Visual perception determines that the socket is aligned with the screw, meeting the engagement requirement. \\
  \texttt{socketed} & Force perception confirms the socket successfully engages with the screw, meeting the disassembly requirement. \\
  \bottomrule 
  \end{tabularx} 
  \end{center}
\end{table}
  
\begin{table}[t]
  \caption{Detailed Description of Action Primitives}
  \label{tab:primitives_description}
  \begin{center}
  \begin{tabularx}{\linewidth}{cX} 
  \toprule  
  Primitive & \multicolumn{1}{c}{Function Description} \\
  \midrule  
  \texttt{Move} & Moves the robotic arm to position the end-effector near the top of the target screw. \\
  \texttt{Mate\_vision} & Moves the arm above the screw, estimates the precise pose via visual perception, and repositions the end-effector to align its center axis with the screw. \\
  \texttt{Mate\_force} & Estimates the precise screw pose through force feedback and moves the arm to align the end-effector's axis with the screw. \\
  \texttt{Insert} & Lowers the end-effector along its axis while rotating to engage the socket with the screw. \\
  \texttt{Disassemble} & Rotates the end-effector counterclockwise to unscrew and disassemble the target screw. \\
  \bottomrule 
  \end{tabularx} 
  \end{center}
\end{table}

\begin{table}[t]
  \caption{PDDL Representation of Action Primitives: Preconditions and Expected Effects}
  \label{tab:PDDL_representation}
  \begin{center}
  \begin{tabularx}{\linewidth}{
  >{\centering\arraybackslash}p{1.6cm}  
  X                                   
}
  \toprule  
  Primitive & \multicolumn{1}{c}{Preconditions and Expected Effects} \\
  \midrule  
  \texttt{Move} &  
  :pre \enspace \texttt{have\_coarse\_pose}~$\wedge$~$\neg$~\texttt{near\_screw} \\
  & :eff \enspace \texttt{near\_screw}~$\wedge$~\texttt{above\_screw} \\  
  \texttt{Mate\_vision} &  
  :pre \enspace $\neg$~\texttt{pattern}~$\wedge$~\texttt{near\_screw}~$\wedge$~$\neg$~\texttt{target\_aim} \\
  & :eff \enspace \texttt{pattern}~$\wedge$~\texttt{above\_screw}~$\wedge$~\texttt{target\_aim} \\
  \texttt{Mate\_force} &
  :pre \enspace $\begin{aligned}[t]
  &\texttt{pattern} \wedge \texttt{near\_screw} \wedge \neg\texttt{above\_screw} \\
  &\wedge \neg\texttt{target\_aim}
  \end{aligned}$\\
  & :eff \enspace $\neg$~\texttt{pattern}~$\wedge$~\texttt{above\_screw}~$\wedge$~\texttt{target\_aim} \\  
  \texttt{Insert} &  
  :pre \enspace \texttt{target\_aim}~$\wedge$~\texttt{above\_screw} \\
  & :eff \enspace \texttt{socketed}~$\wedge$~$\neg$~\texttt{above\_screw} \\ 
  \texttt{Disassemble} &  
  :pre \enspace \texttt{socketed}~$\wedge$~$\neg$~\texttt{disassembled} \\
  & :eff \enspace \texttt{disassembled} \\
  \bottomrule 
  \end{tabularx} 
  \end{center}
\end{table}

We formulate the system-level task as a planning problem defined by a tuple \((S_0, S_G, A)\), where \(S_0\) denotes the initial state, \(S_G\) the goal state, and \(A = \{a_1, a_2, \ldots, a_n\}\) the set of action primitives, with \(a_i\) \((i = 1, \ldots, n)\) representing individual action primitives. The objective of the planning algorithm is to find an executable sequence of action primitives \(\texttt{prim\_list} = \{a_{p_1}, a_{p_2}, \ldots, a_{p_m}\}\) such that applying these actions in order transitions the system from \(S_0\) to \(S_G\).

In this work, we consider the disassembly of electric vehicle (EV) battery screws as a representative application scenario. While this task is trivial for humans, autonomous execution in unstructured environments remains highly challenging for robotic systems. To enable effective planning and execution, multimodal sensory inputs—comprising visual and force data—are processed by neural networks to infer quasi-symbolic state representations in the form of neural predicates (Table~\ref{tab:predicates_description}).

Inspired by human disassembly behavior, we design five core action primitives (Table~\ref{tab:primitives_description}). Each primitive is defined in a PDDL-like formalism using preconditions and expected effects grounded in the neural predicate space (Table~\ref{tab:PDDL_representation}), enabling integration with symbolic task planning frameworks.

\section{METHOD} 
 \begin{figure*}[thpb]
  \centering
  \includegraphics[width=\textwidth]{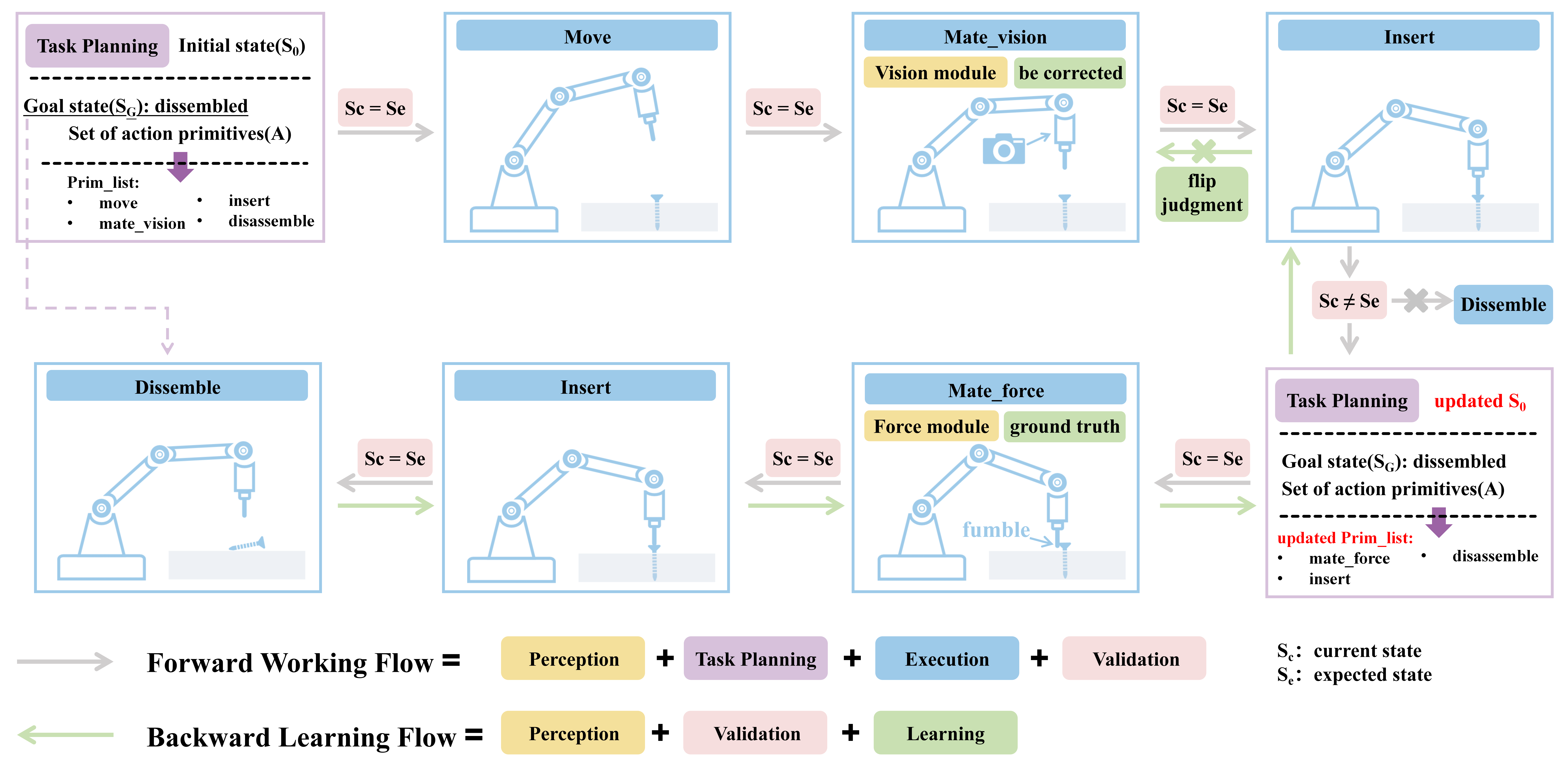}
  \caption{\textbf{Illustration of the bidirectional reasoning flow, consisting of a forward working flow and a backward learning flow.} In the forward working flow, the system generates an action primitive sequence based on the initial and goal states, and sequentially executes operations such as \texttt{Move}, \texttt{Mate\_vision}, and \texttt{Insert}. When the \texttt{Insert} action fails to achieve the expected \texttt{socketed} state, the system triggers replanning, switches to force perception to re-estimate the screw pose, and successfully completes the remaining task. After task completion, the backward learning flow retrospectively analyzes the execution process. The successful screw pose estimated by force perception is treated as ground truth to correct the biased visual perception result. Meanwhile, the system identifies and flips the neural predicate with the lowest confidence among the preconditions of the \texttt{Insert} action, and then terminates the traceback.
}
  \label{fig:ReasoningFlow}
\end{figure*}

\subsection{Methodology Overview}
We propose a continual learning framework built upon a Neuro-Symbolic TAMP architecture. The framework integrates a multimodal perception cross-validation mechanism and establishes a bidirectional reasoning flow, composed of a forward working flow and a backward learning flow. It is designed to enable robots to robustly execute tasks in dynamic environments while continuously accumulating knowledge from execution feedback and optimizing internal models, thereby enhancing the system's autonomous learning and adaptive capabilities.

\subsection{Multimodal Perception Cross-validation Mechanism}
During robotic task execution, a single perception modality is often vulnerable to environmental noise, leading to unstable perception results and consequently affecting the accuracy and robustness of task performance. To enhance the system's adaptability in dynamic environments, we introduce a multimodal perception cross-validation mechanism, which dynamically corrects perception errors and optimizes action strategies through complementary information between modalities.

For clarity, the two perception modalities involved in this mechanism are denoted as modality A and modality B. The system first relies on modality A to estimate the precise pose of the target object (e.g., a screw) and executes the corresponding sequence of planned action primitives. After a critical action (such as \texttt{Insert}) is performed, the system invokes modality B to sense and update the current system state, determining whether it aligns with the expected reasoning outcome. If the state is consistent, the system proceeds to subsequent actions (e.g., \texttt{Disassemble}); otherwise, it triggers replanning. Modality B is then used to re-estimate the target pose, and the updated action sequence is executed, enabling a closed-loop correction between perception and planning.

In this work, we select vision and force as two representative perception modalities. The visual modality uses RGB-D images as input, while the force modality captures variations in force signals during exploratory contact. The system fuses these multimodal observations to update neural predicates and generate accurate pose predictions for the screw.

\subsection{Forward Working Flow}
The forward working flow models the system's task execution logic and serves as the core process for planning and verification. In this flow, multimodal perception data are mapped into interpretable symbolic states through neural predicates. The system formalizes both the current and goal states using PDDL, where each action primitive is defined by its preconditions and expected effects.

Given the initial state $S_0$, the goal state $S_G$, and a set of action primitives $A$, the PDDL planner generates an executable sequence of primitives, denoted as $\texttt{prim\_list}$. The robot attempts to execute this sequence in order. Before executing each primitive, the system verifies whether the current symbolic state satisfies its preconditions using real-time multimodal sensing. If the condition holds, the primitive is executed; otherwise, the current state is updated as a new start state $S_0'$, and the planner replans a new sequence $\texttt{prim\_list}'$ based on $(S_0', S_G, A)$.

Throughout the execution, the multimodal perception cross-validation mechanism continuously monitors the state to ensure consistency with the expected symbolic transitions. This enables dynamic correction of the action plan and robust adaptation to environmental changes or sensing inaccuracies.

\subsection{Backward Learning Flow}
The backward learning flow enables continual self-improvement by tracing execution failures and correcting perception and symbolic reasoning errors. Unlike the forward working flow, which focuses on task planning and execution, the backward learning flow focuses on learning from mistakes and optimizing model parameters.

After each task execution, the system retrospectively analyzes the forward working flow. Two main correction mechanisms are applied: perception estimation correction and neural predicate correction.

\subsubsection{Perception Estimation Correction}
When a task is successfully completed, the final perception result (e.g., screw pose) is treated as ground truth, which is assumed to be obtained from the more reliable perception modality A. If, during execution, the system previously used another modality B to produce a biased estimation, the ground truth can be used to supervise and update the model of modality B.

Let the true pose from modality A be:
\begin{equation}
\mathbf{p}_{\text{true}} = [x, y, z, \theta]
\end{equation}
The estimated pose from modality B is given by:
\begin{equation}
\mathbf{p}_B = f_B(I_B; \theta_B)
\end{equation}
where $f_B$ is the model for modality B, with input $I_B$ and parameters $\theta_B$. We minimize:
\begin{equation}
L_B = \frac{1}{2} \lVert \mathbf{p}_{\text{true}} - \mathbf{p}_B \rVert_2^2
\end{equation}
and update:
\begin{equation}
\theta_B^{t+1} = \theta_B^t - \eta \nabla_{\theta_B} L_B.
\end{equation}

\subsubsection{Neural Predicate Correction}
When an action fails to achieve its expected effects, the system backtracks through its preconditions to identify and correct unreliable neural predicates. Assuming that non-neural predicates are reliable, let the set of $m$ neural predicates involved in the preconditions be denoted as $\{P_1, P_2, \dots, P_m\}$. Each predicate $P_i$ is predicted by a neural network classifier $g_i(x; \phi_i)$, where $x$ represents the input perceptual data and $\phi_i$ denotes the classifier parameters. The classifier outputs a probability distribution:
\begin{equation}
g_i(x; \phi_i) = [p_i^0, p_i^1]
\end{equation}
where $p_i^0$ and $p_i^1$ represent the probabilities that predicate $P_i$ is false (0) and true (1), respectively. The confidence score is defined as:
\begin{equation}
c_i = \max(p_i^0, p_i^1)
\end{equation}
The system identifies the neural predicate $P_k$ with the lowest confidence score based on the outputs of the neural network classifiers:
\begin{equation}
k = \arg\min_{i=1,\dots,m} c_i.
\end{equation}
The prediction for $P_k$ is then flipped to produce corrected label $\hat{y}_k$, and parameters $\phi_k$ of its classifier $g_k$ are updated by minimizing cross-entropy:
\begin{equation}
\hat{y}_k = 1 - y_k, \quad y_k = \arg\max(p_k^0, p_k^1).
\end{equation}
\begin{equation}
L_k = -[\hat{y}_k \log(p_k^1) + (1 - \hat{y}_k)\log(p_k^0)]
\end{equation}
\begin{equation}
\phi_k^{t+1} = \phi_k^t - \eta \nabla_{\phi_k} L_k.
\end{equation}

The process stops after a single correction to avoid introducing unreliable supervision signals into the training set.

\subsection{Illustrative Case}
Figure~\ref{fig:ReasoningFlow} illustrates the application of the proposed bidirectional reasoning flow in an electric vehicle battery screw removal task, comprising both the forward working flow and the backward learning flow.

In this task, the forward working flow begins with the initial state $S_0 = \neg\texttt{pattern} \allowbreak \wedge \texttt{have\_coarse\_pose(pose)} \allowbreak \wedge \neg\texttt{near\_screw}$, the goal state $S_G =$\texttt{disassembled}, and a set of action primitives $A = \{\texttt{Move}, $\allowbreak$ \texttt{Mate\_vision}, $\allowbreak$ \texttt{Mate\_force}, $\allowbreak$ \texttt{Insert}, $\allowbreak$ \texttt{Disassemble}\}$. Based on these inputs, the PDDL planner generates an initial sequence of action primitives $\texttt{prim\_list} = \{\texttt{Move}, $\allowbreak$ \texttt{Mate\_vision}, $\allowbreak$ \texttt{Insert}, $\allowbreak$ \texttt{Disassemble}\}$. The robot executes this sequence in order. During \texttt{Mate\_vision}, the system estimates the screw pose using visual perception. Before executing the \texttt{Insert} action, the system evaluates its two neural predicate preconditions, \texttt{target\_aim} and \texttt{above\_screw}, as true, and proceeds with execution.

However, the \texttt{Insert} action fails, as the force perception module indicates that the screw is not successfully socketed, i.e., the expected \texttt{socketed} predicate is not achieved. The forward working flow then triggers an abnormal state handler, halting the current execution. Based on the updated state $S_0' = $\allowbreak$ \texttt{pattern} $\allowbreak$ \wedge \texttt{near\_screw} $\allowbreak$ \wedge \neg\texttt{above\_screw} $\allowbreak$ \wedge \neg\texttt{target\_aim}$, the planner generates a new action sequence $\texttt{prim\_list'} = \{\texttt{Mate\_force}, $\allowbreak$ \texttt{Insert}, $\allowbreak$ \texttt{Disassemble}\}$. The robot executes the new sequence in order. During \texttt{Mate\_force}, the system re-estimates the screw pose using force perception, eventually completing the task successfully.

After the forward working flow concludes, the backward learning flow retrospectively analyzes the execution. Since the task is successfully completed based on force perception, the resulting screw pose is regarded as ground truth. The system detects that a biased visual estimation was used earlier in the execution. Hence, the system corrects the visual estimation using the ground truth and stores the corrected result along with its corresponding input sample in the continual learning dataset.

Furthermore, because the initial execution of \texttt{Insert} failed, the system infers that some of its preconditions may have been misjudged. Upon review, it identifies that among the neural predicate preconditions, \texttt{target\_aim} has the lowest confidence score. Thus, this predicate is assumed to be incorrectly classified, and its label is flipped. The corrected label and its corresponding input are then stored in the continual learning dataset.

Once a sufficient volume of such corrected samples accumulates in the dataset, the system performs supervised updates of the corresponding model parameters. This process improves the accuracy of perception and the adaptability of the system in dynamic environments.

\section{EXPERIMENTS AND RESULTS}

\subsection{Experimental System Architecture}

\begin{figure}[t]
  \centering
  \includegraphics[width=\linewidth]{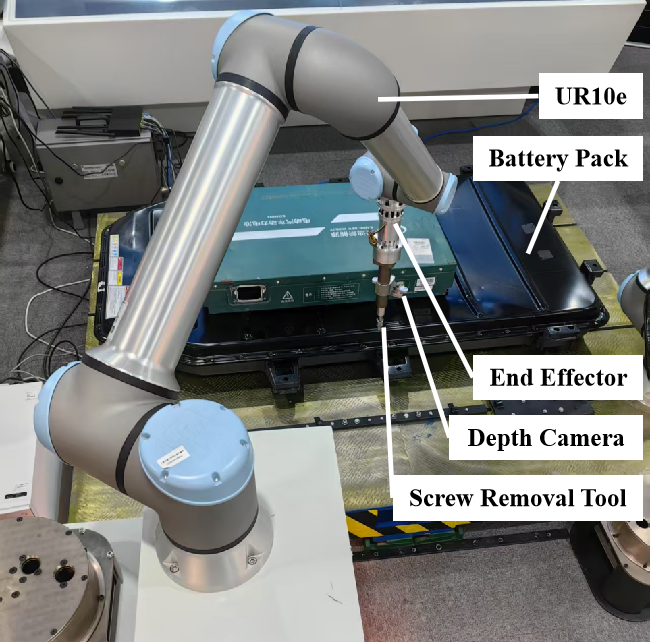}
  \caption{Experimental System Architecture}
  \label{fig:ExpPlatform}
\end{figure}

Figure~\ref{fig:ExpPlatform} shows the experimental system architecture, including the following core components:
\begin{itemize}
    \item Retired onboard battery pack as the disassembly target.
    \item UR10e six-degree-of-freedom collaborative robotic arm.
    \item End-effector integration module:
    \begin{itemize}
        \item Custom screw removal tool.
        \item Intel RealSense depth camera for visual sensing.
    \end{itemize}
    \item ROS-based hybrid symbolic-reactive control system.
\end{itemize}

\subsection{Experimental Setup}
To evaluate the robustness and adaptability of the proposed Neuro-Symbolic TAMP continual learning framework in dynamic and uncertain environments, we design a series of battery screw disassembly tasks that integrate multimodal perception and symbolic planning. Each task aims to successfully remove a single screw. The system deploys both visual perception networks and force perception networks to perform cross-validation and complementary correction.

The perception module updates the predicates defined in Table~\ref{tab:predicates_description} based on either visual or force feedback, which in turn guides the PDDL-based symbolic planner in selecting action primitives (see Table~\ref{tab:primitives_description}). The experimental scenario is designed to reflect realistic operational conditions, particularly lighting variation, which greatly affects the accuracy of visual predictions while having limited impact on force perception.

\paragraph{Replanning Strategy:}
During task execution, if an action (e.g., \texttt{Insert}) fails to achieve its expected effect (e.g., the target state \texttt{socketed} is not satisfied), the system triggers a replanning mechanism. It regenerates an updated sequence of action primitives based on the new state in an attempt to recover task execution.

To prevent excessive recovery attempts due to perception errors, we define a replanning threshold $n_{th} = 10$. When the number of replans $n$ exceeds this threshold, the task is marked as failed, and the system proceeds directly to the next task without further replanning.

\paragraph{Perception Mismatch and Backward Correction:}
For tasks that are eventually successful but require at least one replan ($0 < n \leq n_{th}$), the system initiates a backward learning flow to retrospectively correct perception errors:
\begin{enumerate}
    \item \textbf{Cross-modal Correction:} The final successful pose estimation (typically based on force perception) is regarded as ground truth. This ground truth is used to correct earlier pose predictions from another modality (e.g., vision). The corrected label, along with its corresponding input data, is stored in the continual learning buffer.

    \item \textbf{Neural Predicate Correction:} The system backtracks through the action sequence and identifies the first failed action. Among its neural preconditions, the one with the lowest confidence is considered erroneous and its output is flipped to form a corrected training label. This corrected sample is also stored in the continual learning buffer.
\end{enumerate}

\paragraph{Continual Learning Trigger:}
When the number of newly collected samples in the buffer of any perception module reaches a predefined threshold (e.g., 75), the system triggers an incremental model update. The update process is autonomous and does not interrupt task execution, enabling continual improvement of the perception accuracy.

\subsection{Evaluation Metrics}

We define the number of replans $n$ as a key metric to assess the accuracy of the perception-to-symbol mapping. A replan indicates that the system's perception is insufficient to support the successful completion of the current task. If $n$ exceeds the threshold $n_{th}$, the task is considered a failure. The binary success indicator function for each task is defined as:
\begin{equation}
    \mathcal{I}(n) = 
    \begin{cases}
        1, & n < n_{th} \quad (\text{Task Success}) \\
        0, & n \geq n_{th} \quad (\text{Task Failure})
    \end{cases}
    \label{eq:indicator}
\end{equation}

\paragraph{Quantitative Evaluation Metrics:}
To quantitatively evaluate the performance improvement brought by the continual learning mechanism, we record the replanning count $n$ and calculate the following two metrics after each round of model update:

\begin{itemize}
  \item \textbf{Task Success Rate:}
  \begin{equation}
  \mathrm{SUS} = \frac{1}{N}\sum_{k=1}^{N} \mathcal{I}(n_k)
  \end{equation}
  which reflects the proportion of tasks that are successfully completed within the replanning threshold.
  
  \item \textbf{Average Number of Replans:}
  \begin{equation}
  \bar{n} = \frac{1}{N}\sum_{k=1}^{N} n_k
  \end{equation}
  which measures the system's average reliance on replanning and indicates the stability of perception.
\end{itemize}

\subsection{Results and Discussion} 

\begin{table}[t]
  \caption{Task Success Rate and Average Replans}
  \label{tab:results}
  \centering
  \begin{tabularx}{0.8\linewidth}{>{\centering\arraybackslash}X >{\centering\arraybackslash}X >{\centering\arraybackslash}X >{\centering\arraybackslash}X} 
  \toprule  
  Iterations & Task Count & $\mathrm{SUS}$ & $\bar{n}$ \\
  \midrule  
  0 & 131 & 81.68\%  & 3.389 \\
  1 & 162 & 89.51\%  & 2.648 \\
  2 & 62 & 100.00\%  & 1.984 \\
  3 & 39 & 100.00\%  & 1.128 \\
  \bottomrule 
  \end{tabularx} 
\end{table}

Table~\ref{tab:results} presents the task success rate $\mathrm{SUS}$ and average number of replans $\bar{n}$ over four system iterations. The number of iterations reflects the number of continual learning updates applied to the neural perception modules.

The results demonstrate that the continual learning mechanism enables the system to refine neural-symbolic mappings by leveraging real task outcomes. With each update, the system becomes more reliable: $\mathrm{SUS}$ rises to 100\% after two updates, and $\bar{n}$ reduces by over 60\%. This illustrates improved perception confidence and fewer symbolic mismatches.

In particular, the number of perception-induced replans drops significantly due to better prediction from corrected neural predicates. The system learns to recover from early-stage misjudgments (e.g., misestimated \texttt{target\_aim}) and reduces dependence on replanning. Moreover, experiments show that integrating multi-modal corrections improves robustness to environmental disturbances such as lighting variations.

These findings confirm that our proposed bidirectional reasoning flow, which combines symbolic error recovery and continual perception refinement, significantly enhances task success and reduces planning instability in unstructured robotic manipulation.

\section{CONCLUSIONS}
In this paper, a neural-symbolic TAMP framework integrating multimodal perceptual cross-validation and continual learning is proposed to address the dynamic adaptation challenges of autonomous screw disassembly of power batteries in unstructured scenarios. By constructing a forward working flow and a backward learning flow based on PDDL, the system achieves dynamic correction of perceptual bias and online optimization of symbolic knowledge, which solves the problem of perceptual-symbolic mapping error and performance degradation under environmental perturbation in the traditional method. Experimental validation shows that the proposed framework significantly improves task success rates in dynamic disassembly scenarios (increasing from 81.68\% to 100\%) and reduces the number of perception misjudgments (decreasing from 3.389 to 1.128). 

However, to ensure the smooth interaction between the forward working flow and the backward learning flow, the current system still relies on carefully designed predicates and action primitives, which can pose a considerable engineering burden in complex tasks. Moreover, the backward learning flow assumes that action execution is error-free and that state transitions are fully traceable—assumptions that often do not hold in real-world unstructured environments. Furthermore, during the perception correction process, the system selects the neural predicate with the lowest confidence as the most likely misjudgment for label correction. However, low confidence does not necessarily indicate an incorrect prediction, which may lead to incorrect supervision and degrade the quality of model updates. In addition, when all perception modalities are simultaneously affected by noise or interference, the system may be unable to obtain reliable ground-truth estimations from any single modality, making it difficult to correct the others. These limitations reduce the effectiveness of continual learning and compromise the system's robustness and adaptability in complex environments.

In future work, we plan to incorporate action success rates and state transition probabilities into the framework to enable more robust probabilistic reasoning. We also aim to explore the integration of continual learning with adaptive action primitive generation based on large language models (LLMs). By leveraging LLMs' knowledge and generalization abilities, the system can dynamically generate and refine action primitives during execution, adapting to novel environments. LLMs can further assist in defining new symbolic operations and optimizing primitive parameters through multimodal feedback, enabling more autonomous skill acquisition and improving the system's generalization and decision-making capabilities in complex scenarios.

\addtolength{\textheight}{-12cm}   


\end{CJK*}

\bibliographystyle{unsrt} 
\bibliography{main}

\end{document}